\documentclass{article}

\usepackage{spconf,amsmath,graphicx}

\usepackage{times}          
\usepackage{amsmath}        
\usepackage{amssymb}        
\usepackage[table]{xcolor}  
\usepackage{hyperref}      

\usepackage{algorithm}      
\usepackage{algpseudocode}  
\usepackage{enumitem}      
\usepackage{multirow}       

\title{Agent-Driven Large Language Models for Mandarin Lyric Generation}
\name{Hong-Hsiang Liu, Yi-Wen Liu} 
\address{Author Affiliation(s)}
\address{National Tsing Hua University \\
  }
\begin{document}
%
\maketitle
\begin{abstract}
Generative Large Language Models have shown impressive in-context learning abilities, performing well across various tasks with just a prompt. Previous melody-to-lyric research has been limited by scarce high-quality aligned data and unclear standard for creativeness. Most efforts focused on general themes or emotions, which are less valuable given current language model capabilities. In tonal contour languages like Mandarin, pitch contours are influenced by both melody and tone, leading to variations in lyric-melody fit. Our study, validated by the Mpop600 dataset, confirms that lyricists and melody writers consider this fit during their composition process. In this research, we developed a multi-agent system that decomposes the melody-to-lyric task into sub-tasks, with each agent controlling rhyme, syllable count, lyric-melody alignment, and consistency. Listening tests were conducted via a diffusion-based singing voice synthesizer to evaluate the quality of lyrics generated by different agent groups.

\end{abstract}
\section{Introduction}
\label{sec:intro}


The release of ChatGPT by OpenAI in 2022 has highlighted the extensive applications of Large Language Models (LLMs). Based on GPT-3, ChatGPT facilitates tasks such as conversation, question answering, translation, and summarization without retraining\cite{brown_language_2020}. 
Compared to traditional Seq2Seq models\cite{seq2seq}, the GPT model, leveraging the Transformer architecture\cite{Transformer}, excels in contextual understanding. Consequently, tasks like lyric generation and creative writing have transitioned to GPT models.


In traditional machine learning, ample labeled data is essential for training. However, lyric generation tasks often lack sufficient aligned melody and lyric data due to constraints like copyright and expertise \cite{tian_unsupervised_2023}, leading to the use of plain text data \cite{tsaptsinos2017,bejan2020, edmonds2021}. Transformer-based GPT models, though trained on extensive textual data, convert text into tokens without incorporating phonetic and prosodic information \cite{Transformer}. Therefore, it is not trivial to prompt them into considering melodic information when generating lyrics for a song.

Phonetic alignment with melody is crucial in lyric writing. Redesigning the data acquisition and training pipeline could enable large language models to learn music-related information, but the high parameter count and training costs make this impractical \cite{brown_language_2020}. Also, traditional machine learning tasks typically have standard answers, allowing models to be trained these via loss functions. However, lyric creation is subjective and lacks standard answers, making traditional methods unsuitable.

Despite its subjectivity, lyric creation has constraints. Given the same melody, different lyrics may vary in how they can be sung and understood with ease. Tian et al.~proposed an unsupervised inference method for English lyric generation, using a checking mechanism to ensure generated lyrics conform to the melody \cite{tian_unsupervised_2023}. This study extends their method to Chinese lyric generation.

\section{Data Format and Problem Definition}
\label{sec:problem_definition}

The melody of a song is composed of numerous notes and can be parsed into multiple phrases consisting of consecutive notes \cite{tian_unsupervised_2023}. Therefore, the melody $M$ of a song can be defined as:

\begin{equation}
    M = \{ p_1, p_2, ...\},
    \label{eq:melody}
\end{equation}
where $p_i (i = 1, 2, ...)$ represents a phrase, and each phrase is a sequence of notes:
\begin{equation}
    p_i = \{ n_{i1}, n_{i2}, ..., n_{iN_i} \}.
    \label{eq:phrase}
\end{equation}
Further, each note $n_{ij} (j = 1, 2, ..., N_i)$ can be represented as a combination of a pitch value $f_{ij}$ and a duration $d_{ij}$. Thus, a note can be denoted as  $n_{ij} = (f_{ij}, d_{ij})$, where $f_{ij} \in \{0, 1, ..., 127\}$ and $d_{ij} \in \mathbb{R^+}$.
Here, we represent a rest by $f_{ij} = 0$ while $f_{ij} \neq 0$ corresponds a specific key on the semitone scale from low to high.

Referring to previous studies \cite{Lyric_gen_4, Lyric_gen_3, tian_unsupervised_2023}, the melody-to-lyric (M2L) problem can be defined as follows: Given a phrase $p_i$, generate the lyrics $L_i$ that match the musical melody. Here, $L_i$ can be represented as:
\begin{equation}
    L_i = \{ w_{i1}, w_{i2}, ..., w_{iN_i} \},
    \label{eq:lyric}
\end{equation}
where $w_{ij} (j = 1, 2, ..., N_i)$ represents a Chinese character or a space (corresponding to a rest), and $N_i$ denotes the number of characters corresponding to the phrase $p_i$.

\section{Related Work}
\label{sec:related_work}

\subsection{Lyric Generation}
\label{subsec:lyric}
Earlier studies on lyric generation relied on manually predefined rules, often resulting in lyrics that lacked musicality \cite{oliveira2007tralalyrics} \cite{oliveira2015tralalyrics20}. With advancements in neural networks and deep learning, data-driven lyric generation models have emerged \cite{Lyric_gen_1} \cite{potash2015ghostwriter}. 
Research in this area has included emotion-controlled lyric generation, such as the mLSTM-based model by Ferreira et al. \cite{Lyric_gen_by_emotion_control}, and theme-controlled lyric generation, like the model proposed by Chang et al. \cite{chang_singability-enhanced_2021}. 
Given the capabilities of large language models \cite{brown_language_2020}, controlling themes or emotions without considering melody has become a simpler task.

\subsection{Mandarin Lyrics}
\label{subsec:mandarin}

Melody-lyric alignment has been carefully considered in automatic English lyric generation \cite{chang_singability-enhanced_2021}. Building on phenomena identified by Dzhambazov et al.~\cite{realtion_between_lyric_and_melody_in_english}, 
Tian et al.~\cite{tian_unsupervised_2023} explored the rules governing the correspondence between English lyrics and melody. They concluded that duration has a greater impact than pitch in English songs. Consequently, they proposed a method that aligns long notes with stressed syllables and short notes with unstressed syllables to generate lyrics that better fit the musical rhythm.

Studies on Chinese lyric generation, however, are relatively scarce. 
Unlike English and most other spoken languages, Modern Mandarin is a tonal language and it primarily has four tones: Yinping (tone 1), Yangping (tone 2), Shang (tone 3), and Qu (tone 4). 
Thus, conflicts may arise if the tonal direction of the lyrics differs from the melody's pitch direction \cite{relation_between_lyric_and_melody_in_mandarin}. This mismatch is known as \emph{inverted tones} or \emph{mismatched lyrics and melody}, while a good match is referred to as \emph{aligned tones}, or matched lyrics and melody \cite{Xuefan}.

To reduce the conflict between lyrics and melody, Xue \cite{Xuefan} proposed some rules to help lyricists align lyrics with melody. Sun et al.~\cite{Sun1988} suggested suitable pitch relationships for the four Mandarin tones, ranked from high to low: Tone 1  > Tone 4  > Tone 2  > Tone 3. These theories converge on similar principles, as summarized in Table \ref{tab:melody_direction}, which outlines suitable melodic directions for sequences of Mandarin characters with varying tones. This alignment is especially relevant when multiple characters are present in a phrase; for single characters, any pitch is generally acceptable.

\begin{table}
  \renewcommand{\arraystretch}{1.5}
  \centering
  \begin{tabular}{|c|>{\centering\arraybackslash}p{1cm}|>{\centering\arraybackslash}p{1cm}|>{\centering\arraybackslash}p{1cm}|>{\centering\arraybackslash}p{1cm}|}
   \hline
   \rowcolor{gray!30} Previous/Next& Tone1 & Tone2 & Tone3 & Tone4 \\
   \hline
   Tone1 & \cellcolor{blue!30}Desc & \cellcolor{blue!30}Desc & \cellcolor{blue!30}Desc & \cellcolor{blue!30}Desc \\
   \hline
   Tone2 & \cellcolor{red!30}Asc & \cellcolor{blue!30}Desc & \cellcolor{blue!30}Desc & \cellcolor{red!30}Asc \\
   \hline
   Tone3 & \cellcolor{red!30}Asc & \cellcolor{red!30}Asc & \cellcolor{blue!30}Desc & \cellcolor{red!30}Asc \\
   \hline
   Tone4 & \cellcolor{red!30}Asc & \cellcolor{blue!30}Desc & \cellcolor{blue!30}Desc & \cellcolor{blue!30}Desc \\
   \hline
  \end{tabular}
  \caption{Melodic directions and their suitability to different two-character tone combinations. Desc = descending, and Asc = ascending.}
  \label{tab:melody_direction}
\end{table}

There has been considerable research on the applicability of such rules in music. Ling found that mismatched lyrics and melody increased cognitive errors in 4-year-old native Chinese-speaking children \cite{Tone_match_for_child_song}. Wee analyzed 10 Chinese songs and found that 649 out of 668 syllables adhered to lyric-melody alignment rules \cite{relation_between_lyric_and_melody_in_mandarin}. Ho discovered that 80.3\% of cases in popular songs aligned diagonally, indicating a significant correlation between the tone of the lyrics and the direction of the melody \cite{Ho_2006}.

\subsection{Prompt Engineering and Agent Methods}
\label{subsec:prompt_engineering}

In the early stages of prompt engineering, strategies focused on making tasks more comprehensible for LLMs, such as the Chain of Thought (CoT) approach \cite{CoT_2022}, which breaks down complex problems into smaller steps, and the Self-Reflection method \cite{Self_reflection_LLM}, which encourages models to reflect on previous outputs to improve robustness.
Strategies have since evolved to involve multiple agents working collaboratively. For example, the Exchange of Thought (EoT) method divides tasks among agents to sequentially think and exchange results \cite{EoT_2023}, thus enhancing reasoning and robustness. Similarly, the Multi-Agent Debate approach involves models debating from different perspectives to improve reasoning \cite{Multi_agent_debeat}.

Given that LLMs primarily excel at text generation but struggle with complex tasks requiring precise operations, researchers have proposed methods like Retrieval-Augmented Generation (RAG) and tool-using agents to augment LLM capabilities. RAG retrieves and references external databases before generating responses, while tool-using agents enhance LLM capabilities by providing external tools and necessary parameters.

In this study, we adopt tool-using agents by equipping different agents with specific tools and organizing them in a particular sequence. This approach aims to achieve rhyme control, syllable count control, and alignment between lyrics and melody, thereby enhancing the overall capability of LLMs in Mandarin lyric composition.

\section{Methods}
\label{sec:method}

\subsection{Forward Control and Backward Control for LLMs}
\label{subsec:control}

\begin{figure}
  \centering
  \includegraphics[width=0.5\textwidth]{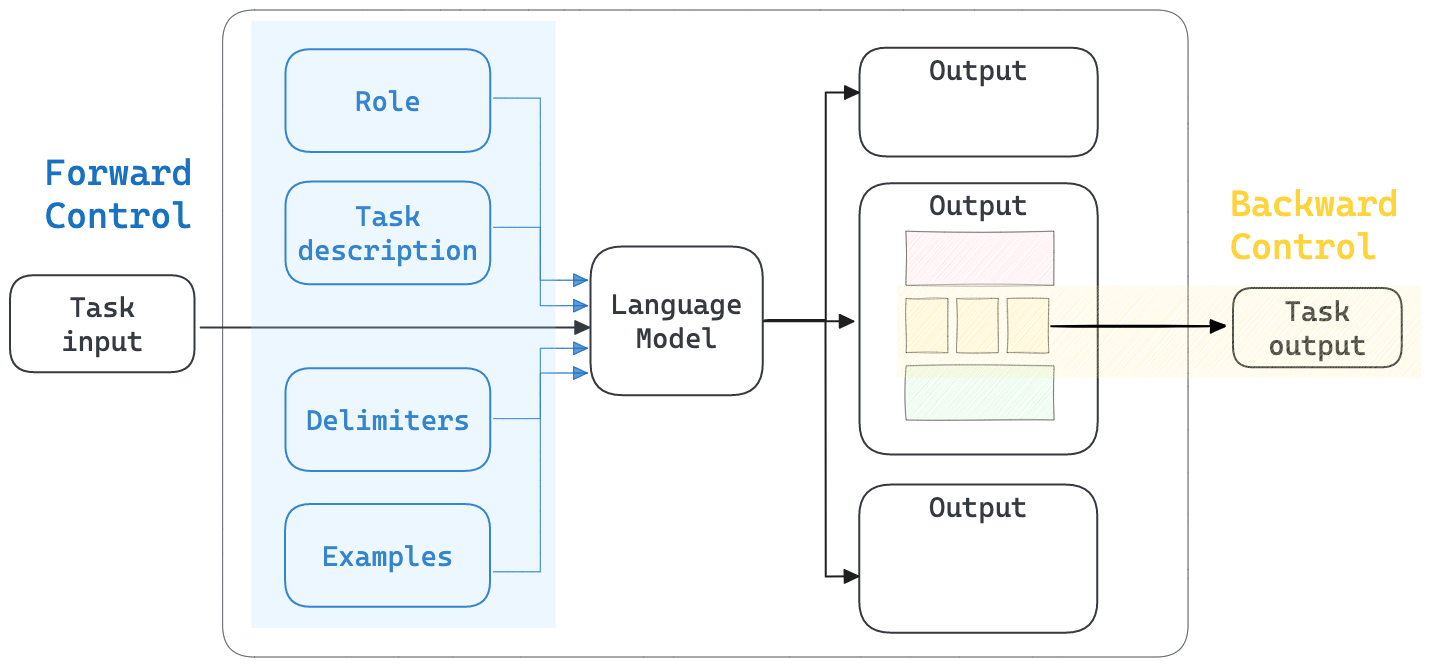}
  \caption{Forward Control and Backward Control for LLMs}
  \label{fig:forward_backward_control}
\end{figure}

This study employs a GPT-based language model for lyric generation, which outputs tokens. To ensure the generated lyrics meet specific format requirements, control mechanisms are necessary. As illustrated in Figure \ref{fig:forward_backward_control}, these mechanisms can be categorized into \emph{forward control}, which adjusts the input fed into the model, and \emph{backward control}, which processes the model's output to influence the final content.

In the forward control section, prompts are used to guide the model. According to the prompt writing guidelines provided by OpenAI \cite{openai2024prompt}, effective prompts can be crafted by clearly describing the task objectives, assigning a specific persona to the model, using delimiters to distinguish different input parts, and providing concrete examples for reference.
The prompts used in this study's agents are crafted based on these strategies. Each prompt includes a persona suitable for the task, a clear task objective description, and delimiters to separate task requirements and reference materials. Additionally, examples are provided to enhance the model's output quality. 

\begin{figure}[htb]
  \centering
  \includegraphics[width=0.35\textwidth]{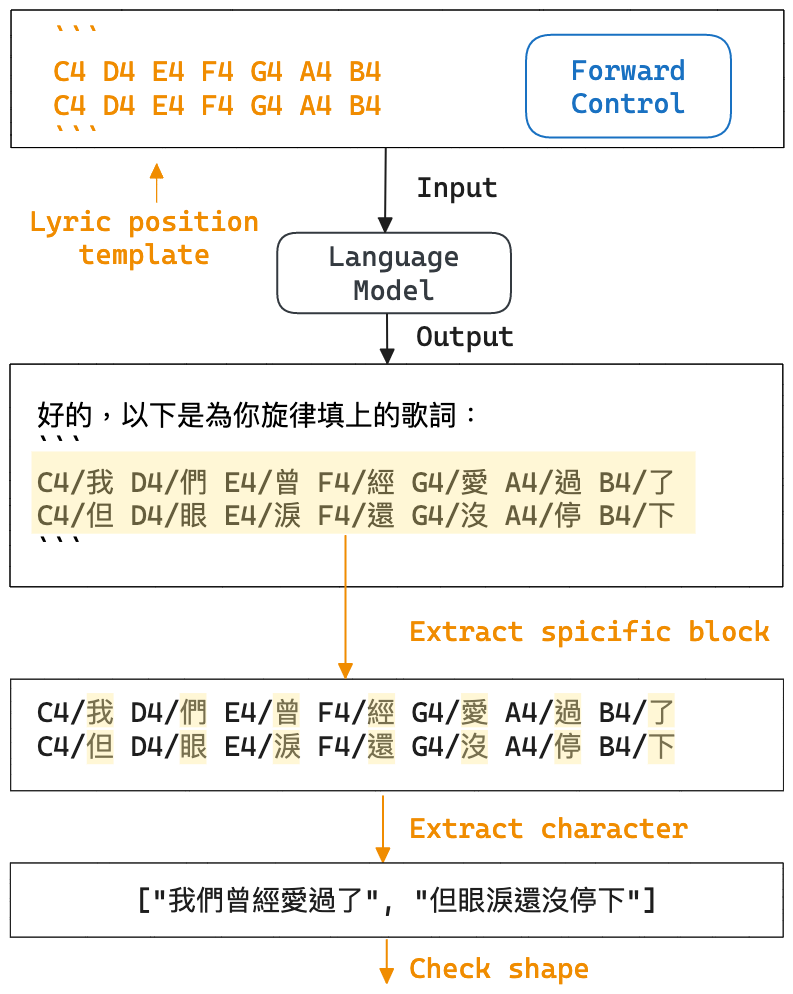}
  \caption{Backward control process}
  \label{fig:backward_control}
\end{figure}

Despite providing clear task objectives and output examples, the model may still produce content that deviates from the desired format. Therefore, post-processing of the model's output is necessary for better control. As shown in Figure \ref{fig:backward_control}, the backward control process involves:

\begin{itemize}[noitemsep]
  \item Providing reference grids for lyrics creation to guide the model.
  \item Extracting specific formatted blocks from responses to avoid extraneous content.
  \item Combining extracted blocks into the final lyrics.
  \item Further checking the final lyrics to ensure they meet format requirements.
\end{itemize}

\subsection{The Proposed Agent Method}

As illustrated in Algorithm \ref{alg:lyric_generation}, the generation of lyrics $L$ for a song involves sequentially extracting segments {\color{blue}$M_i$} of the melody $M$ and feeding them into the agent group $G$ for generation. Optional additional requirements 
$R$ can be provided.

\begin{algorithm}
  \small 
    \caption{Segment-by-Segment Lyric Generation}
    \label{alg:lyric_generation}
    \begin{algorithmic}[1]
    \Require Melody $M$ of a song and optional additional requirements $R$
    \Ensure Lyrics $L$ of the song
    \Procedure{M2L}{$M, R$}
        \State $L \gets [\ ]$ \Comment{Initialize lyrics}
        \State $G \gets [g_1, \cdots, g_N]$ \Comment{Initialize agent group}
        \For{each $M_i$ in $M$} \Comment{Generate lyrics segment by segment}
            \State $m \gets M_i$ \Comment{Extract structure of the melody}
            \State $l \gets G(m, L, R)$ \Comment{Generate lyrics using the agent group} 
            \State $L \gets L + [l]$ \Comment{Append generated lyrics to $L$}
        \EndFor
        \State \textbf{return} $L$
    \EndProcedure
    \end{algorithmic}
\end{algorithm}

\begin{figure}[htb]
  \centering
  \includegraphics[width=0.4\textwidth]{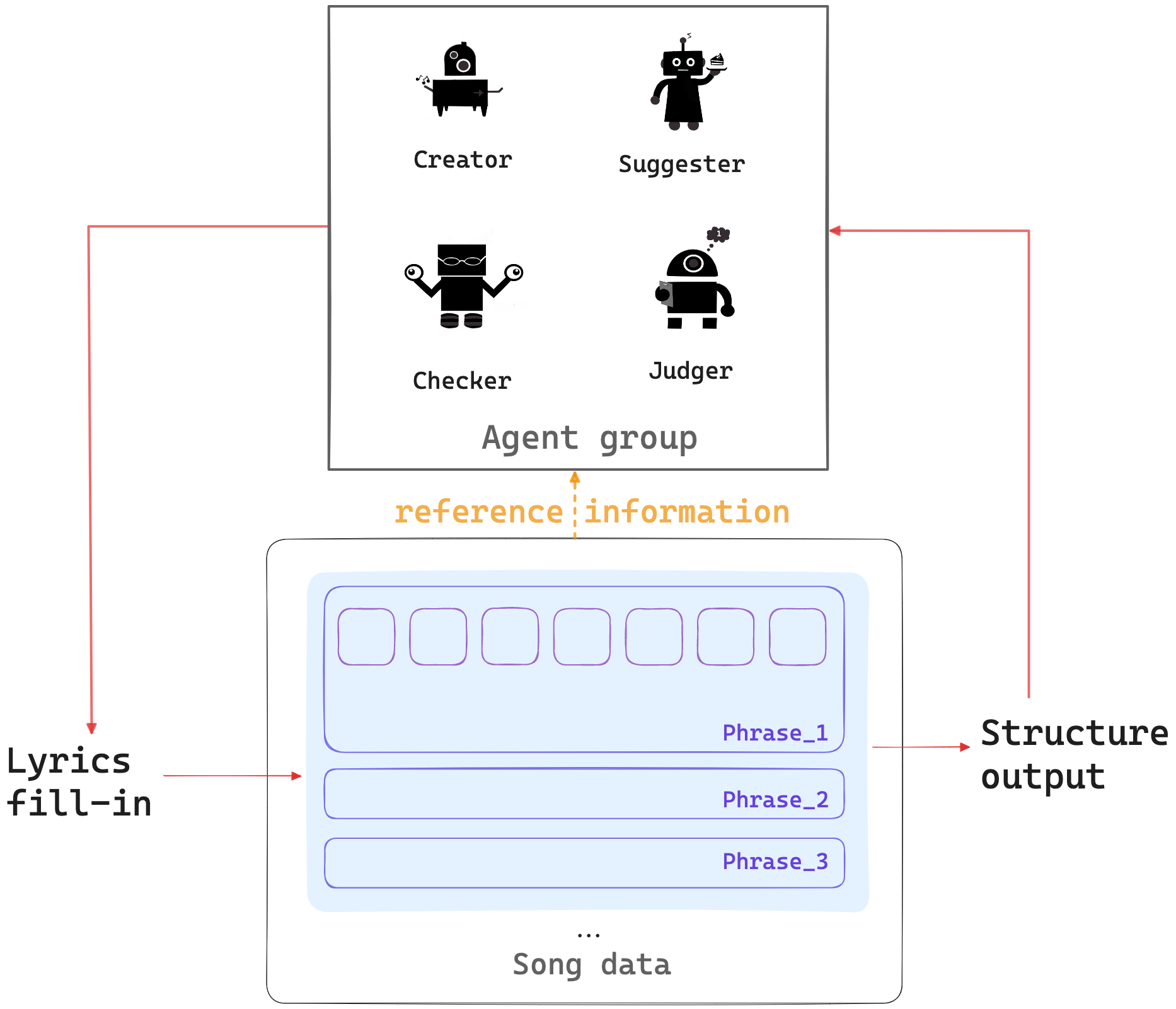}
  \caption{The flow of lyric generation by angents}
  \label{fig:lyric_generation_flow}
\end{figure}

In each generation step involving the agent group, up to four agents can collaborate to generate a segment of lyrics. These agents are:

\begin{enumerate}
    \item Suggester: Providing rhyming suggestions based on the preceding lyrics.
    \item Creator: Generating the corresponding lyrics based on the position in the melody, taking into account the rhyming suggestions from the previous agent.
    \item Checker: Examining the cohesion between the generated lyrics and the melody, thus providing further suggestions.
    \item Judger: Making the final selection from multiple possible options based on the generated lyrics and cohesion suggestions, and requesting a re-generation if necessary.
\end{enumerate}
Their roles are further explained below.

\paragraph*{Suggester}


Due to the inherent limitation that language models process input as text tokens without tonal information, the Suggester first analyzes the preceding lyrics to extract appropriate words. These words are then matched with external rhyme dictionaries. The results are formatted and provided to the Creator to facilitate subsequent lyric generation, allowing for controlled rhyming in the produced lyrics.

\paragraph*{Creator}


Drawing on the rhyming words provided by the preceding Agent, the Creator maps these words to appropriate positions based on the pitch information from the Melody. The method described in \ref{subsec:control} is employed to regulate the number of characters, resulting in lyrics of a specific length.

\paragraph*{Checker}


The Checker performs two primary tasks on the lyrics generated by the Creator. First, it compares the lyrics against the melody using the rules outlined in Table \ref{tab:melody_direction} to assess the alignment between lyrics and melody, marking any discrepancies in terms of the number and type of mismatched words. Second, it employs a language model to compare the generated lyrics with the preceding ones, evaluating contextual consistency and providing relevant feedback.

\paragraph*{Judger}


In the final stage, the Judger reviews each set of lyrics generated by the Creator along with the corresponding feedback from the Checker. Based on this information, the Judger selects the most suitable lyrics. If none of the options meet the required criteria, the Judger may request a regeneration, initiating the lyric creation process anew.





\section{Experiment and Results}

\subsection{Importance of Lyric-Melody Alignment in Lyric Creation}

\begin{figure}[htb]

  \begin{minipage}[htb]{\linewidth}
    \centering
    \centerline{\includegraphics[width=8cm]{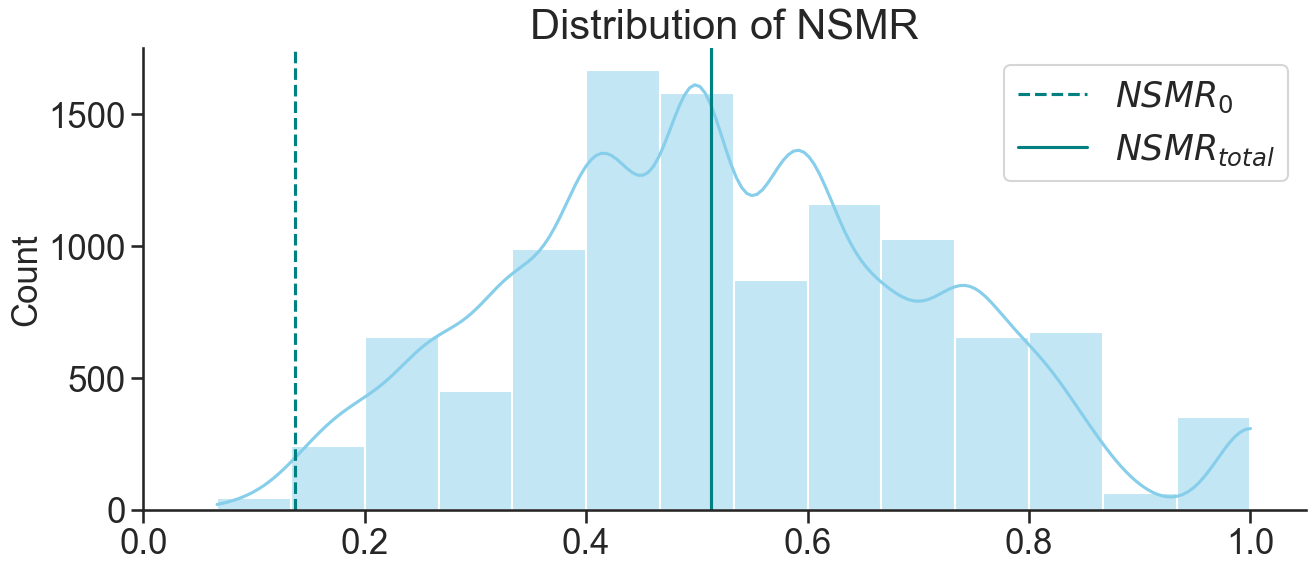}}
  \end{minipage}
  \\[0.2cm]
  \begin{minipage}[htb]{\linewidth}
    \centering
    \centerline{\includegraphics[width=8cm]{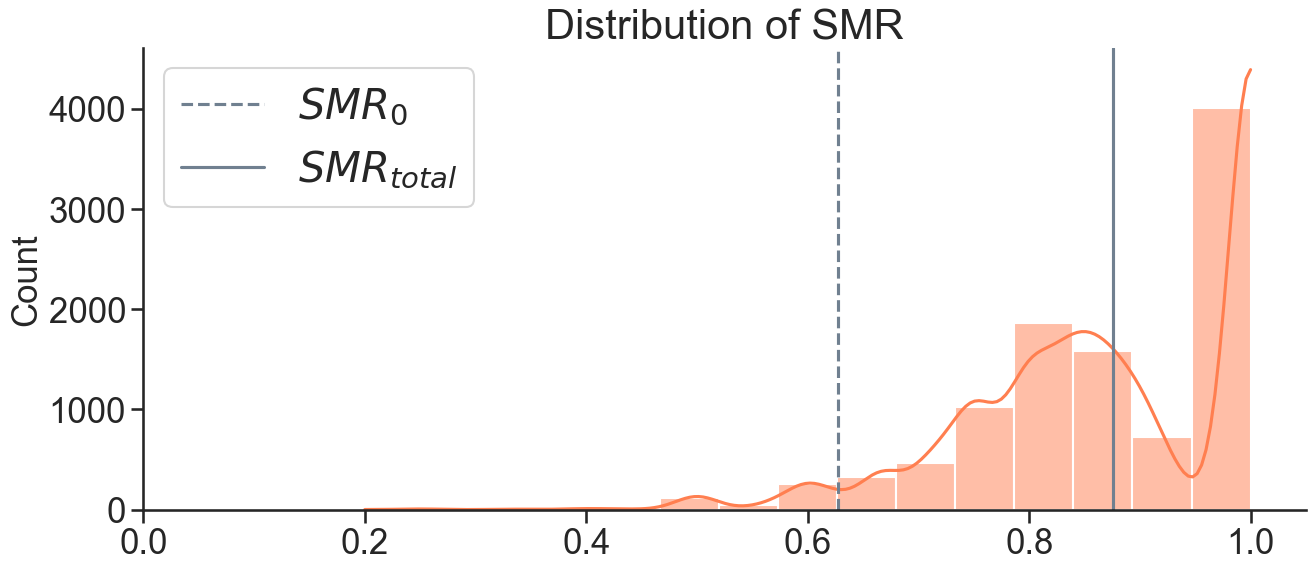}}
  \end{minipage}
  \caption{Distribution of NSMR and SMR in Mpop600.}
  \label{fig:exp1-NSMR-SMR}
\end{figure}

In this experiment, we conducted a statistical analysis on a total of 10,427 phrases from songs in the Mpop600 dataset \cite{Chu2020MPop600}. The results were divided into two groups based on whether Chinese word segmentation (Word Segment) was considered: Segment Match (SM) and Non-Segment Match (NSM). By dividing the number of matched characters 
by the total number of characters, we obtained the Non-Segment Match Rate (NSMR) and Segment Match Rate (SMR). The distribution of these rates is shown in Figure \ref{fig:exp1-NSMR-SMR}. The overall NSMR and SMR were 51.3\% and 87.6\%, respectively. 

The first character of each phrase and the first character of each word will inevitably match in NSM and SM, respectively. If lyricists were to completely contradict the melody-lyrics alignment rules, that would lead to minimum values, $\text{NSMR}_{0}$ and $\text{SMR}_{0}$, at 13.7\% and 62.7\%, respectively.
Therefore, we can expect the NSMR and SMR to be approximately $(\text{NSMR}_{0} + 1 )/ 2$ and $(\text{SMR}_{0} + 1 )/ 2$ under completely random conditions, which are 56.85\% and 81.35\%, respectively. Note that the actual $\text{NSMR}$ is lower than that of random conditions, 
while the actual $\text{SMR}$ is higher than that of the random condition. These results indicate that, during the process of Mandarin lyric creation, the importance of phonetic alignment within the same word is higher than that of arbitrary adjacent characters within the same phrase.



\subsection{Chatacter Count Control}

In this experiment, we aim to control language models to generate content with a specific character count by combining different methods and models. We will compare the following approaches:

\begin{itemize}[noitemsep]
  \item \textbf{Unrestricted}: Directly instructing the language model to generate content with a specific character count, serving as the baseline for comparison.
  \item \textbf{Prompting}: Guiding the language model to generate content with a specific character count by adding prompts during the generation process.
  \item \textbf{Formatting and Examples}: Incorporating formatting and examples to guide the language model in generating content with a specific character count.
  \item \textbf{Fill-in-the-Blank}: Providing fill-in-the-blank templates to facilitate the generation of content with a specific character count.
\end{itemize}

For each method, we used different models and requested various character counts for comparison. The models used in this experiment included Gemini-1.0-pro, gpt-3.5-turbo-0125, and gpt-4-0125-preview. For different groups, we requested the generation of content with 5, 10, and 20 Mandarin characters. Each combination of group and character count underwent 10 trials, and we calculated the exact character count accuracy rate.

\begin{table}
  \begin{center}
    \small 
    \renewcommand{\arraystretch}{1.2}
    \begin{tabular}{|c|c|c|c|}
        \hline
        \textbf{Method} & \textbf{Gemini} & \textbf{gpt-3.5} & \textbf{gpt-4} \\
        & \textbf{-1.0-pro} & \textbf{-turbo-0125} & \textbf{-0125-preview} \\
        \hline
        \textbf{Unrestricted} & 27\% & 13\% & 7\% \\
        \hline
        \textbf{Prompting} & 23\% & 13\% & 3\% \\
        \hline
        \textbf{Formatting} & 20\% & 23\% & 10\% \\
        \hline
        \textbf{Formatting} & \multirow{2}{*}{23\%} & \multirow{2}{*}{43\%} & \multirow{2}{*}{33\%} \\
        \textbf{and Example}& & & \\
        \hline
        \textbf{Fill in} & 0\% & 43\% & \textbf{80\%} \\
        \hline
    \end{tabular}
    \caption{Exact character count accuracy rates for different methods and models}
    \label{tab:exp2-accuracy}
  \end{center}
\end{table}


Table \ref{tab:exp2-accuracy} presents the character count accuracy rates for five methods using three different models. The results from the Unrestricted group indicate that the exact character count accuracy was low across the various language models. No significant improvement was observed in the Prompting and Formatting groups. However, the combination of Formatting and Example methods led to substantial improvements in both GPT models. Finally, in the Fill-in group, a notable improvement was achieved with the gpt-4-0125-preview model, reaching an 80\% accuracy rate for exact character counts.

\subsection{Listening Test}


To assess whether the number of agents impacts the quality of generated lyrics, we randomly selected three popular Chinese songs and employed four distinct agent groups to generate lyrics for each song. Subsequently, 22 participants were invited to partake in a listening test, in which they hear the songs rendered by a diffusion-based singing voice synthesizer \cite{svs_yinping}. Then, participants were asked to rank four versions of each song, presented in randomized orders. The rankings provided by all participants were then averaged to derive the mean ranking for each of the four agent groups across the three selected songs. The four agent groups were as follows:
\begin{itemize}[noitemsep]
  \item \textbf{Group 1}: Creator
  \item \textbf{Group 2}: Creator + Judger
  \item \textbf{Group 3}: Creator + Checker + Judger
  \item \textbf{Group 4}: Suggester + Creator + Checker + Judger
\end{itemize}

\begin{figure}[htb]
  \centering
  \includegraphics[width=0.4\textwidth]{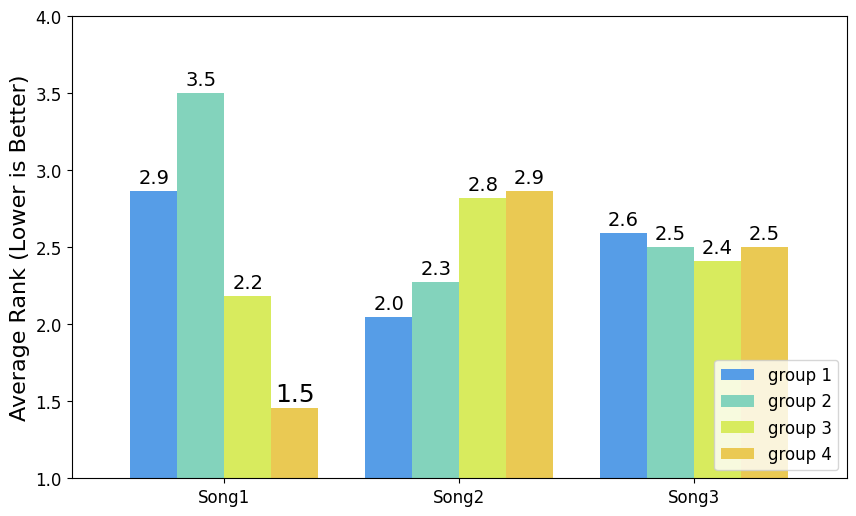}
  \caption{Average rank of different agent groups, as reported by 22 listeners (lower is better)}
  \label{fig:exp3-average_rank}
\end{figure}


From Figure \ref{fig:exp3-average_rank}, it can be observed that increasing the number of agents does not necessarily guarantee an improvement in the quality of the generated lyrics. We speculate that this may be due to the inherent randomness in the lyric generation process. However, it is noteworthy that for one of the songs, Group 4 exhibited a significant deviation in average ranking compared to the other groups. Specifically, 68\% of participants assigned it the highest rank, with none assigning it the lowest rank. This result suggests that, under certain circumstances, employing a larger number of agents might increase the upper limit of the quality of generated lyrics. Further experiments may be necessary to substantiate this finding.


\section{Conclusion}





In this research, we attempt to use LLMs for Mandarin lyrics generation.
Besides adjusting the input prompt to the language model, we also propose a method called backward control, which performs post-processing to the language model output. In our experiments, the backward control method enabled GPT-4 to achieve an 80\% accuracy rate in generating Chinese lyrics with the exact specified number of characters.Finally, we propose a framework that integrates multiple agents to enhance different aspects of lyric generation, including rhyme control, character count control, lyric-melody alignment, and consistency control. Based on this multi-agent architecture, we supply the agents with various external tools to extend the capabilities of the language models. By recruiting 22 subjects to listening to lyrics synthesized by different agent groups, we observe that while the use of a larger number of agents does not guarantee superior results in every instance.Nevertheless, in one of the three melodies, employing all of the agents clearly creates the best lyrics.
Future research shall explore the criteria for the present system to work optimally and formulate different strategies for the agents to collaborate in the Mandarin lyrics writing task.




\bibliographystyle{IEEEbib}
\bibliography{refs}

\end{document}